\documentclass[letterpaper, 10 pt, conference]{ieeeconf}  % Comment this line out if you need a4paper

\IEEEoverridecommandlockouts                              % This command is only needed if 
\usepackage{times}
\usepackage{epsfig}
\usepackage{graphicx}
\usepackage{amsmath}
\usepackage{amssymb}
\usepackage{graphicx}
\usepackage{amsmath}
\usepackage{booktabs}
\usepackage{algorithm}
\usepackage{algorithmic}
\usepackage{comment}
\usepackage{bm} 
\usepackage{tikz}
\usepackage{xcolor}         % colors
\usepackage{booktabs}
\usepackage{multirow}
\usepackage{colortbl}
\usepackage{wrapfig}
\usetikzlibrary{arrows.meta,positioning,fit,backgrounds,calc,shadows.blur}

\title{\LARGE \bf
Adaptive Event Stream Slicing for Open-Vocabulary Event-Based Object Detection via Vision-Language Knowledge Distillation 
}

\author{Jinchang Zhang, Zijun Li, Jiakai Lin,  Guoyu Lu
\thanks{Jinchang Zhang, Zijun Li, Jiakai Lin and Guoyu Lu are with the Intelligent Vision and Sensing (IVS) Lab at SUNY Binghamton, USA. 
        {\tt\small guoyulu62@gmail.com}}%
}
\begin{document}

\maketitle
\thispagestyle{empty}
\pagestyle{empty}

%%%%%%%%%%%%%%%%%%%%%%%%%%%%%%%%%%%%%%%%%%%%%%%%%%%%%%%%%%%%%%%%%%%%%%%%%%%%%%%%
\begin{abstract}
Event camera offers advantages in object detection tasks for its properties such as high-speed response, low latency, and robustness to motion blur. However, event cameras inherently lack texture and color information, making open-vocabulary detection particularly challenging. Current event-based detection methods are typically trained on predefined target categories, limiting their ability to generalize to novel objects, where encountering previously unseen objects is common.
Vision-language models (VLMs) have enabled open-vocabulary object detection in RGB images. However, the modality gap between images and event streams makes it ineffective to directly transfer CLIP to event data, as CLIP was not designed for event streams.
To bridge this gap, we propose an event-image knowledge distillation framework, leveraging CLIP’s semantic understanding to achieve open-vocabulary object detection on event data. Instead of training CLIP directly on event streams, we use image frames as teacher model inputs, guiding the event-based student model to learn CLIP’s rich visual representations. Through spatial attention-based distillation, the student network learns meaningful visual features directly from raw event inputs, while inheriting CLIP’s broad visual knowledge.
Furthermore, to prevent information loss due to event data segmentation, we design a hybrid Spiking Neural Network (SNN) and Convolutional Neural Network (CNN) framework. Unlike fixed-group event segmentation methods, which often discard crucial temporal information, our SNN adaptively determines the optimal event segmentation moments, ensuring that key temporal features are extracted. The extracted event features are then processed by CNNs for object detection.
\end{abstract}

\section{Introduction}
\begin{figure*}[t]
\begin{center}
%\framebox[4.0in]{$\;$}
\includegraphics[width=17cm, height=7.5cm]{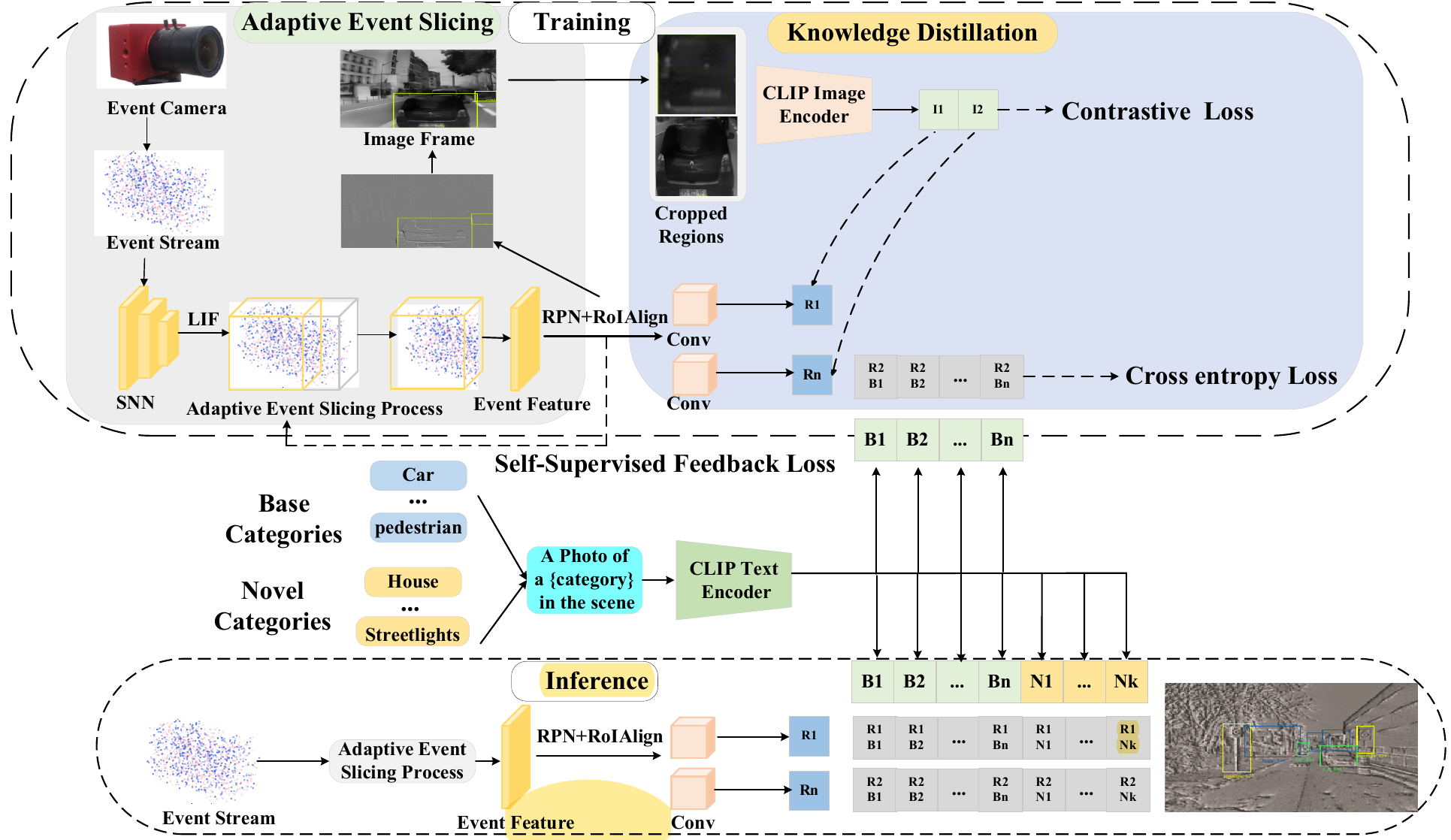}
\end{center}
\vspace{-5 mm}
\caption{The overview of our framework. The event stream is first fed into a spiking neural network, where Self-Supervised Feedback Loss is utilized to dynamically adjust the membrane potential based on object detection results, enabling adaptive event segmentation and feature extraction. We transfer image knowledge from CLIP to event data, using the CLIP image encoder as a teacher model. Through knowledge distillation, the student detector trained on event data learns the rich visual representations from CLIP. Additionally, category text is input into the frozen CLIP text encoder to generate text embeddings, and the cosine similarity between each region embedding and all category text embeddings is computed for object classification.
During the inference phase, the model performs open-vocabulary object detection using only event stream data, without relying on image frames. }
\vspace{-6mm}
\label{overallframework}
%\vspace{-3 mm}
\end{figure*}

Event cameras  are bio-inspired vision sensors that fundamentally differ from traditional frame-based cameras. They capture event streams asynchronously and sparsely, gaining attention for their superior characteristics, including high temporal resolution, high dynamic range, low latency, and low power consumption. In recent years, leveraging these inherent advantages, event-based vision perception has advanced across various domains, including object tracking, depth estimation, object detection. 
%数据集缺失，模态不对的问题

As one of the core tasks in event-based perception, event-based object detection has gained significant attention but remains constrained to closed-set settings. Due to the unique imaging modality of event cameras and the lack of large datasets, existing models struggle to generalize to unseen categories in real-world scenarios. Consequently, open vocabulary object detection for event cameras has become a critical challenge.
In frame-based vision tasks, pretrained vision-language models (VLMs) such as CLIP \cite{radford2021learning} have achieved open vocabulary detection by learning aligned image-text representations from large datasets. %While leveraging the VLMs for open vocabulary event camera detection appears promising, direct application leads to severe overfitting, where models still tend to detect only previously seen categories. 
However, CLIP or other VLM \cite{xu2025affordance} was designed for regular visible images, instead of event camera streams. This issue primarily stems from the modality gap between event data and conventional 2D images, making CLIP inapplicable to raw event streams. Moreover, the absence of large-scale event-text datasets makes it impractical to train a vision-language model for event cameras from scratch.
Another challenge of event cameras lies in slicing of the raw event stream.
When using event streams as input, two key steps are required: (1) segmenting  the raw event stream into multiple sub-event groups and (2) converting these sub-event groups into different event representations. Current research focuses on optimizing event representations \cite{wang2019event}, while overlooking the crucial segmentation step. Common segmentation methods \cite{li2025achieving} employ fixed grouping strategies, such as slicing event streams based on a fixed number of events  or a fixed time interval. However, these approaches suffer from information imbalance—potentially leading to information loss in low-speed motion scenarios and excessive redundancy in high-speed motion conditions. Although some recent studies \cite{peng2023better} have proposed adaptive sampling methods, with \cite{zhang2023frame} requiring explicit searches over multiple time periods, they still do not directly learn an adaptive segmentation process.

To address the event stream slicing problem, we propose the Adaptive Event Slicing module, which leverages spiking neural networks to dynamically determine the optimal event segmentation timing, overcoming the limitations of traditional fixed slicing strategies. Additionally, to bridge the modality gap between event streams and image frames, which prevents the direct application of pretrained VLMs, we introduce Event-Based Vision-Language Knowledge Distillation. This framework employs the image encoder from VLMs as the teacher network and an event-based object detection model as the student network, enabling knowledge transfer from VLMs to event data.
Specifically, during SNN-based event triggering, we adaptively determine the optimal event segmentation timing and extract event stream features for subsequent object detection. To enhance the stability of event segmentation, we introduce the Linear Incremental Constraint Loss, preventing premature spike activations. Additionally, we design the Self-Supervised Feedback Loss (SSF-Loss), which adjusts the membrane potential based on object detection results, guiding the SNN to fire at the optimal time step and adaptively refine the event segmentation strategy.
In the object detection module, we employ a category-agnostic proposal module to improve the model’s ability to generalize to unseen object categories. To align event-based features with image representations, we extract image features from a frozen CLIP image encoder and perform feature alignment with event stream embeddings from the same ROI region. We adopt CLIP’s image encoder as the teacher model and incorporate knowledge distillation based on a spatial self-attention mechanism, enabling the student detector trained on event data to learn rich visual representations from CLIP. Furthermore, category text embeddings are generated by a frozen CLIP text encoder and integrated with region embeddings for object classification.
During the inference phase, we perform open-vocabulary object detection using only event stream data, without relying on image frames.

In summary, our main contributions include the following aspects: 
1. To the best of our knowledge, we are the first to introduce an event-based open vocabulary object detection framework, enabling object detection directly based on textual descriptions.
2. We design a knowledge distillation method leveraging CLIP to enhance event-based object detection, effectively transferring rich semantic knowledge to event data.
3. We propose a self-supervised spiking neural network slicing feedback scheme, which dynamically adjusts the membrane potential according to object detection results, enabling adaptive event segmentation and feature extraction.
Figure \ref{overallframework} shows our framework.

\section{Related Work}

\subsection{Event-based Object Detection}

In event-based object detection, mainstream approaches can be broadly grouped into two families: CNN-based models and energy-efficient, biologically inspired SNN-based methods.
CNN-based methods typically convert event streams into frame-like representations—such as event histograms, time surfaces, and event volumes—so they can be processed by existing deep learning detectors \cite{wang2024airshot,zheng2026blockchain,li2024size,bao2025towards,zhao2024balf,edstedt2024dedode,ni2023feature}. These studies also extend detection beyond RGB image object detection, including\cite{zhang2024underground,lin20253d,lin2025keypoint,ning2025multi,pan2026finscra}
However, these frame-based encodings often discard the intrinsic spatiotemporal information of event data, thereby limiting detection performance. To mitigate this, recent work such as SNN-based methods focus on exploiting the sparsity and computational efficiency of event data. Traditional SNN detectors largely rely on ANN-to-SNN conversion, as in Spiking-YOLO \cite{hu2023fast}. Yet such approaches typically require many timesteps to match ANN performance, which limits real-time applicability \cite{qu2024spiking}. To address this, directly trained SNN detection frameworks have been proposed, including EMS-YOLO and SpikeYOLO \cite{przewlocka2024poweryolo}. EMS-YOLO introduces an all-spiking residual block (EMS-ResNet) \cite{zhou2024deep}, while SpikeYOLO integrates integer-valued LIF (I-LIF) neurons to reduce quantization error \cite{wang2024object}.
Furthermore, recent advances such as SFOD and CREST further optimize SNN-based detection on event-camera data: SFOD leverages multi-scale feature fusion, and CREST introduces a spiking spatiotemporal IoU loss \cite{wang2024object,xu2024serial}.

\subsection{Open-Vocabulary Object Detection}
Open-Vocabulary Object Detection has evolved from zero-shot object detection based on visual attributes to open-vocabulary detection leveraging VLMs \cite{chen2024ov}, progressively enhancing the model's generalization ability to unseen categories.
Early ZSD methods \cite{zhu2024survey} relied on visual attributes for unseen category inference. However, these methods suffer from limited scalability and generalization, making them less effective for large-scale open-vocabulary detection.
With the rise of VLMs researchers have explored replacing detector classifiers with text embeddings and employing cross-modal feature alignment to enhance open-vocabulary detection \cite{du2024lami}. For instance, ViLD leverages knowledge distillation to inject open-vocabulary knowledge into two-stage detectors \cite{gu2021open}. 
PromptDet \cite{feng2022promptdet} enhances vision-text embedding alignment using learnable prompts.
Recent research has focused on cross-modal contrastive learning and spatiotemporal feature fusion to adapt event data for VLMs. EventCLIP \cite{wu2023eventclip} converts event data into 2D grids and employs an adapter to align event features with CLIP knowledge. E-CLIP \cite{shin2022clip} adopts Hierarchical Triple Contrastive Alignment, jointly optimizing event, image, and text embeddings.However, open-vocabulary detection models trained on images perform poorly when directly applied to event cameras due to the modality gap. 
Therefore, for heterogeneous modalities, it is crucial to design and select a suitable distillation scheme for knowledge transfer \cite{li2025frequency}.
To address this issue, we propose a knowledge transfer method to adapt the knowledge learned from image-based models to event-based detection models.
%写两句上面方法存在的问题，我们来解决

\vspace{-2mm}

\section{Adaptive Event stream Slicing Process}
\vspace{-2mm}
An event stream is an asynchronous data representation, defined as a set:
$E = \{ [x_i, y_i, t_i, p_i] \}_{i=1}^{N}$
with a temporal span of \( T \), i.e., \( t_i \in [t_0, t_0 + T] \). To convert the event stream into a format suitable for Spiking Neural Networks (SNNs), we adopt the voxel grid representation method \cite{bardow2016simultaneous}. Given that SNNs naturally align with event stream data, we utilize them as event stream slicers to enable a dynamic event slicing process and enhance object detection performance.
For an event stream, slicing is determined by the state of spiking neurons (excited/resting). As dynamic event triggers, spiking neurons execute event slicing upon generating a spike at the current time step \( n_c \), where 
$S_{\text{out}} = 1$
indicates a slicing event. This operation captures the time interval from the last spike to the current one, forming an event group within this interval, which can then be used for object detection tasks.

\textbf{Membrane Potential Driven Loss:}
Since the event segmentation position depends on whether SNN generates a spike at a given time step, it is essential to guide the network to precisely trigger a spike at the optimal time step \( n^* \). Specifically, if the segmentation is expected to occur at time \( n^* \), the neuron's membrane potential should reach the threshold \( V_{th} \) at this moment to generate a spike. However, since the membrane potential resets to its resting state immediately after a spike, this process may introduce inaccuracies in guiding subsequent time steps \cite{cao2024spiking}. 
 Therefore, we impose supervision on the non-reset membrane potential \( U[n] \) at \( n^* \), ensuring that \( U[n^*] \geq V_{th} \). The Membrane Potential Driven Loss (Mem-Loss) as follows:
 \vspace{-2mm}
\begin{equation}
\mathcal{L}_{\text{Mem}} = \left\lVert U[n^*] - (1 + \alpha)V_{th} \right\rVert_2^2,
\vspace{-2mm}
\end{equation}
\noindent where \( \alpha \geq 0 \) is a hyperparameter that controls the extent to which the expected membrane potential surpasses the threshold. This loss function effectively guides the spiking neuron to fire at the target time step during training, thereby enabling precise segmentation of the event stream.

\textbf{Linear Incremental Constraint Loss:}
Due to the hill effect \cite{cao2024spiking}, if the membrane potential at a later time step is guided above the threshold, an earlier time step may trigger a spike first, causing the neuron to enter a resting state and thereby affecting the accuracy of event segmentation. Even with the introduction of the membrane potential-driven loss (Mem-Loss), premature triggering of segmentation time steps may still occur. To ensure stable spike triggering, we expect the membrane potential at later time steps to monotonically increase, reducing the impact of premature activation and improving the robustness of segmentation time steps.
To this end, we propose a \textbf{Linear Incremental Constraint Loss} to ensure that the membrane potential at the target time step \( n^* \) precisely reaches the threshold. We establish the following linear growth assumption:
\vspace{-3mm}
\begin{equation}  
U[n_c] \geq U[n^*] \cdot \left( \frac{n_c}{n^*} \right)^\beta,
\vspace{-1mm}
\end{equation}
\noindent where \( n^* \) represents the target spike time step, \( n_c \) denotes the current spike time step, the exponential factor \( \beta \) controls the rate of membrane potential growth.
This constraint ensures that the membrane potential maintains an increasing trend across the entire time domain, effectively preventing excessive early activation from suppressing spikes at later time steps, thereby enhancing the temporal consistency of segmentation time steps.
Based on the above constraint, we define the \textbf{Linear Incremental Constraint Loss} as follows:
\vspace{-2mm}
\begin{footnotesize}
\begin{equation}
\mathcal{L}_{\text{LA}} =
\begin{cases}  
\left\| U[n_c] - V_{th} \cdot \left( \frac{n_c}{n^*} \right)^\beta \right\|^2,  & \text{if } U[n_c] \geq U[n^*] \cdot \left( \frac{n_c}{n^*} \right)^\beta; \\  
0, & \text{otherwise}.
\vspace{-2mm}
\end{cases}
\end{equation}
\end{footnotesize}
This loss is designed to ensure that the membrane potential remains monotonically increasing at time step \( n \), preventing premature spike triggering and enhancing temporal consistency. Additionally, it precisely controls the membrane potential at the target time step \( n^* \) to reach the threshold \( V_{th} \), optimizing spike triggering accuracy. The introduction of the exponential factor \( \beta \) allows for adjustable membrane potential growth rates, enabling the model to adapt to different event distributions.

\textbf{Self-Supervised Feedback Loss:}
To enhance the event slicing performance of SNN in object detection tasks, we propose a Self-Supervised Feedback Loss (SSF-Loss). This loss enables SNN’s event slicing points to adaptively optimize the downstream object detection performance, rather than relying solely on a fixed loss function for optimization.
Specifically, we introduce the object detection task loss \( L_M(n) \) as a feedback signal to directly guide the slicing strategy of the SNN. The loss function is defined as follows:
\vspace{-3mm}
\begin{equation}
L_{\text{SSF}} = \sum_{n} \left( L_M(n) \cdot \left| U[n] - V_{th} \right| \right)
\vspace{-2mm}
\end{equation}
%\vspace{-2mm}
\noindent where \( U[n] \) represents the membrane potential of the SNN at time step \( n \), determining whether a spike is triggered at that step; \( V_{th} \) is the threshold for spike firing; and \( L_M(n) \) represents the loss of the downstream object detection task, reflecting whether the current event slicing aids detection.
If a spike at time step \( n \) increases object detection loss, the optimization adjusts the membrane potential \( U[n] \) to lower slicing probability at that step, shifting it toward a more optimal timing. Conversely, if \( L_M(n) \) is low, indicating beneficial slicing, the SNN favors spiking at that step for better event utilization.
This self-supervised mechanism enables the SNN to refine event slicing over time, extracting key events more effectively and enhancing detection performance in dynamic scenes.

\section{Open-Vocabulary Event Object Detection }
\subsection{Localiziation For Novel Categories}

The core challenge of open-vocabulary object detection lies in accurately localizing unseen object categories. To address this, we enhance the standard two-stage object detector (Mask R-CNN \cite{he2017mask}) by introducing a self-supervised slicing SNN feature extraction network to replace the traditional feature extraction module, making it more suitable for handling the sparsity and asynchronicity of event streams. This design effectively captures the dynamic characteristics of event data, improving the detector's performance on event-based inputs.
To further enhance open-vocabulary object detection  capabilities, we adopt a category-agnostic design, where object detection is performed based on visual features and region proposals rather than predefined category labels. Within this framework, we optimize bounding box regression and mask prediction, replacing the category-specific bounding box regression and mask prediction layers with a category-agnostic proposal module.
This module predicts a single generic bounding box and mask for each region of interest (RoI) rather than generating separate predictions for each category. By transitioning from category-specific predictions to category-agnostic predictions, the model significantly improves its generalization ability to unseen object categories, making it more suitable for open-vocabulary detection tasks.

\textbf{Category-Agnostic Bounding Box Regression:}
In category-agnostic bounding box regression, we remove category-specific bounding box prediction, allowing all objects to share a single set of bounding box regression parameters, formulated as:
\vspace{-1mm}
\begin{equation}
    \mathcal{L}_{box}^{CA} = \sum_{i} 1_{\{y_i > 0\}} \sum_{j \in \{x, y, w, h\}} \text{SmoothL1}(b_{ij} - b_{ij}),
\end{equation}
\noindent where \( i \) denotes the index of the detected object;
\( j \in \{x, y, w, h\} \) represents the four bounding box parameters (center coordinates \( x, y \), width \( w \), and height \( h \));
\( 1_{\{y_i > 0\}} \) is an indicator function ensuring that regression is applied only to foreground objects (excluding background);
\( b_{ij} \) represents the predicted bounding box parameters;
\( b_{ij}^* \) represents the ground-truth bounding box parameters;
\( \text{SmoothL1}(\cdot) \) denotes the Smooth L1 loss function.
This optimization removes dependency on class labels, focusing solely on object localization, which improves generalization to unseen categories.

\textbf{Category-Agnostic Mask Prediction} 
Similarly, in category-agnostic mask prediction, we eliminate category-specific mask prediction, ensuring that all objects share a single mask prediction mechanism:
\vspace{-1mm}
\begin{equation}
    \mathcal{L}_{mask}^{CA} = - \sum_{i} M_i^* \log M_i + (1 - M_i^*) \log (1 - M_i),
    \vspace{-2mm}
\end{equation}
\noindent where:
\( i \) denotes the object index;
\( M_i \) represents the predicted object mask, with values in the range \( [0,1] \);
\( M_i^* \) is the ground-truth binary mask, where 1 represents foreground pixels and 0 represents background pixels;
This loss function employs binary cross-entropy loss (BCE Loss) to measure the similarity between the predicted and ground-truth masks.
By eliminating class dependencies in mask prediction, the model becomes more adaptable to unseen object instances, significantly improving generalization capability in open-vocabulary object detection.

\subsection{Image-to-Event Contrastive Distillation}

Once candidate proposals are generated, we leverage a pretrained vision-language model (CLIP) to classify each region, thereby enabling open vocabulary object detection. However, while CLIP excels in frame-based vision tasks, event cameras perceive dynamic scenes in an asynchronous and sparse manner, resulting in a significant modality gap in data distribution and feature representation compared to conventional images. As a result, directly applying CLIP to zero-shot event-based detection leads to high errors.
To bridge this modality gap, we propose cross-modal knowledge distillation, transferring the image-based knowledge from CLIP’s pretrained model to an event-based object detector. Specifically, we use the CLIP image encoder as a teacher model and apply knowledge distillation to enable the event-based student detector to learn CLIP’s rich visual representations, thereby improving its generalization in open vocabulary event-based detection.

We divide the categories in the detection dataset into a base category subset and a novel category subset, denoted as \( C_B \) and \( C_N \), respectively. The model is trained only using annotations from \( C_B \). In the pre-trained CLIP model, the text encoder and image encoder are represented as \( \mathcal{T}(\cdot) \) and \( \mathcal{V}(\cdot) \).
We train a proposal generation network on the base categories \( C_B \) and extract region proposals ${r_e} \in {P_e}$
 from the event stream. Subsequently, we reconstruct the corresponding image frames from the event stream and map the proposals ${r_e}$ onto the image frames  ${P_i}$. These candidate regions ${r_i}$ are then cropped and resized before being fed into the frozen CLIP image encoder \( \mathcal{V} \) to compute the image embedding:
$\mathcal{V}(\text{crop}(P_i, r_i))$.
To transfer the image knowledge from the CLIP pre-trained model to an event-based object detection model, we align the event-region embedding detected from the event stream, \( \mathcal{R}(\phi(P_e), r_e) \), with the image embedding extracted by the CLIP image encoder from the proposal regions in the image frames, \( \mathcal{V}(\text{crop}(P_i, r_i)) \). This process aims to bridge the modality gap between event data and image frames, enabling the effective utilization of CLIP’s pre-trained visual representations.
To further enhance cross-modal feature alignment, we apply a contrastive loss between event camera region embeddings and image frame embeddings to minimize their representation discrepancy. Additionally, we introduce trainable projection layers, which map event features $f_{{r_e}}^{evt}$ and image frame features $\mathcal{V}(\text{crop}(P_i, r_i))$ into the same feature space, ensuring efficient cross-modal knowledge transfer.
% Furthermore, we employ a contrastive learning objective function:
% \begin{equation}
% \mathcal{L}_{F2E}(\theta_e, \omega_e, \omega_f) = - \sum_{i} \log \left[ \frac{e^{\langle f_i^{evt}, \mathcal{V}(\text{crop}(P_i, r_i)) \rangle / \tau_1}}{\sum_{j \neq i} e^{\langle f_i^{evt}, f_j^{img} \rangle / \tau_1}} \right]
% \end{equation}
% \noindent where: \( \langle \cdot, \cdot \rangle \) represents the similarity measure between event embeddings and image embeddings. \( \tau_1 > 0 \) is a temperature coefficient, controlling the contrastive learning strength for positive and negative sample pairs during distillation.
In the process of knowledge distillation for event features, we take into full consideration the sparsity of event data and its prominent edge characteristics. To better extract and align event features with image features, we introduce an enhanced spatial attention mechanism into the distillation process.
Specifically, we first utilize high-level semantic information from the teacher network to generate a spatial attention map, which highlights key information regions in the event data while suppressing redundant noise. The spatial attention map is constructed as follows:
\vspace{-2mm}
\begin{equation}
\mathcal{P}(F)_{i,j} = \frac{1}{C} \sum_{c=1}^{C} |F_{c,i,j}|,
\mathcal{N}(F) = \text{softmax} \left( \frac{\mathcal{P}(F)}{\tau} \right),
\vspace{-2mm}
\end{equation}
where \(\mathcal{P}(F)\) represents the average pooling result along the channel dimension, and \(\mathcal{N}(F)\) is the normalized attention map obtained through softmax, which adjusts the feature distribution to focus on important spatial locations.
Next, we fuse the attention maps of event features \(\mathcal{N}(F^{evt})\) and image features \(\mathcal{N}(F^{img})\) to obtain the final spatially enhanced representation:
$A = \frac{\mathcal{N}(F^{evt}) + \mathcal{N}(F^{img})}{2},$
We then apply a transformation to the event features as \(F^{*} = \mathcal{G}(F^{evt})\), where \(\mathcal{G}\) is a mapping module that adjusts the feature scale.
On this basis, our event-to-image knowledge distillation loss function \(\mathcal{L}_{F2E}\) is modified as follows:
\vspace{-3mm}
\begin{equation}
\mathcal{L}_{F2E}(\theta_e, \omega_e, \omega_f) = - \sum_{i} \log \left[ \frac{e^{\langle A_i \cdot f_i^{evt}, \mathcal{V}(\text{crop}(P_i, r_i)) \rangle / \tau_1}}{\sum_{j \neq i} e^{\langle A_i \cdot f_i^{evt}, f_j^{img} \rangle / \tau_1}} \right],
\end{equation}
where the spatial attention weight \( A_i \) is applied to the event feature \( f_i^{evt} \), ensuring that the event features focus more on key regions, thereby enhancing the distillation effect. Ultimately, this strategy improves the semantic consistency of event features, enabling more effective alignment with image features during the knowledge distillation process and enhancing the model's generalization ability in object detection tasks.
Through this method, we achieve VLM knowledge transfer to event cameras without requiring large-scale event-text dataset training, enabling open vocabulary object detection while mitigating generalization issues caused by the modality gap.

\subsection{Classification For Novel Categories}

For open vocabulary object detection, another critical challenge is how to classify novel category samples. We address this issue by replacing the traditional classifier with text embeddings extracted from CLIP. Specifically, we embed category names into a prompt template (e.g., "A photo of a {class} in the scene.") and pass them through the text encoder \( \mathcal{T} \) to generate category text embeddings.
Our objective is to enable the knowledge-distilled event region features to be classified using text embeddings. During training, we use only \( \mathcal{T}(C_B) \), i.e., the text embeddings of base categories \( C_B \). For proposal regions that do not match any ground-truth annotations in \( C_B \), we classify them as background categories. Since the textual representation "background" may not adequately describe these unmatched proposals, we allow the background category to learn its own embedding \( \mathbf{e}_{bg} \), ensuring it acquires an independent representation in the semantic space.
To achieve this, we compute the cosine similarity between each region embedding \( \mathcal{R}(\phi(P_e), r_e)) \) and all category embeddings, including both \( \mathcal{T}(C_B) \) and \( \mathbf{e}_{bg} \). We apply softmax normalization with a temperature parameter \( \tau \) to compute the \( \mathcal{L}_{CE} \) cross-entropy loss, optimizing the classification distribution. Meanwhile, to train the first-stage region proposal network in the two-stage object detector, we extract region proposals \( r_e \in P_e \) and train the detector, with the loss function defined as: $e_r = \mathcal{R}(\phi(P_e), r_e))$ 
  \vspace{-2mm}
\begin{equation}
    \mathbf{z}(r) = \left[ \text{sim}(e_r, e_{bg}), \text{sim}(e_r, t_1), \dots, \text{sim}(e_r, t_{|C_B|}) \right],
\end{equation}
 \vspace{-4mm}
\begin{equation}
    \mathcal{L}_{\text{text}} = \frac{1}{N} \sum_{r_e  \in P_e } \mathcal{L}_{\text{CE}} \left( \text{softmax}(\mathbf{z}(r_e )/\tau), y_r \right),
    \vspace{-2mm}
\end{equation}

\noindent where \( \text{sim}(\mathbf{a}, \mathbf{b}) = \mathbf{a}^{\top} \mathbf{b} / (\|\mathbf{a}\| \|\mathbf{b}\|) \), 
\( t_i \) denotes elements in \( \mathcal{T}(C_B) \), 
\( y_r \) denotes the class label of region \( r_e  \), 
\( N \) is the number of proposals per event (\(|P_e |\)).

\textbf{Inference:}
During inference, we rely solely on event stream data for open vocabulary object detection, in contrast to the training phase, where image frames from the event stream were additionally used. Meanwhile, we introduce novel categories \( C_N \) to extend the model's open vocabulary detection capability. Our goal is that the knowledge distilled from CLIP’s image representations can enhance generalization of the event-based object detection model to \( C_N \).

\section{Experiment}
\subsection{Datasets and Implementation}
NCAR Dataset: The NCAR dataset \cite{sironi2018hats} is a binary classification dataset consisting of 12,336 car samples and 11,693 background samples. Each sample spans a duration of 100 ms and exhibits varying spatial dimensions.
GEN1 Automotive Detection Dataset: The first-generation automotive detection dataset \cite{de2020large} consists of 39 hours of event camera recordings with a resolution of 304×240. Overall, the dataset contains 228,000 car bounding boxes and 28,000 pedestrian bounding boxes.
The DSEC dataset \cite{gehrig2021dsec} is a high-resolution, large-scale event-frame dataset designed for real-world driving scenarios, captured with a 640×480 resolution event camera alongside RGB image frames. The original dataset lacked object detection annotations, so we utilize the labels introduced in \cite{tomy2022fusing}, which include three object categories: cars, pedestrians, and large vehicles.

\textbf{Implementation}
ViLD with Enhanced Teacher Models.
For experiments with stronger teachers (CLIP ViT-L/14, ALIGN), we adopt EfficientNet-B7 as the backbone and the ViLD-ensemble architecture. RoI features are extracted only from FPN level P3, and the image jittering range is reduced to [0.5, 2.0].
For CLIP ViT-L/14 (768-d embeddings), the fully connected layers in Faster R-CNN heads are expanded to 1,024 dimensions, and the FPN feature dimension is set to 512.
For ALIGN, which combines an EfficientNet-L2 image encoder with a BERT-large text encoder, we modify Mask R-CNN to better distill teacher knowledge. The ViLD-image head is enhanced with EfficientNet MBConvBlocks, followed by global average pooling to produce embeddings consistent with ALIGN. The ViLD-text head keeps the original Faster R-CNN design. Since ALIGN outputs 1,376-d embeddings (~2.7× CLIP), the fully connected layers in the ViLD-text head are expanded to 2,048 units, with FPN features increased to 1,024 dimensions.

% Text Prompts.
% Because the classifier is trained on full sentences, category names are wrapped with prompt templates. We use an ensemble of 63 templates, including several adapted for detection, e.g., “There is {article} {category} in the scene.”

\begin{figure}[t]
%\vspace{-0.1cm}
\begin{center}
\includegraphics[width=8cm, height=5cm]{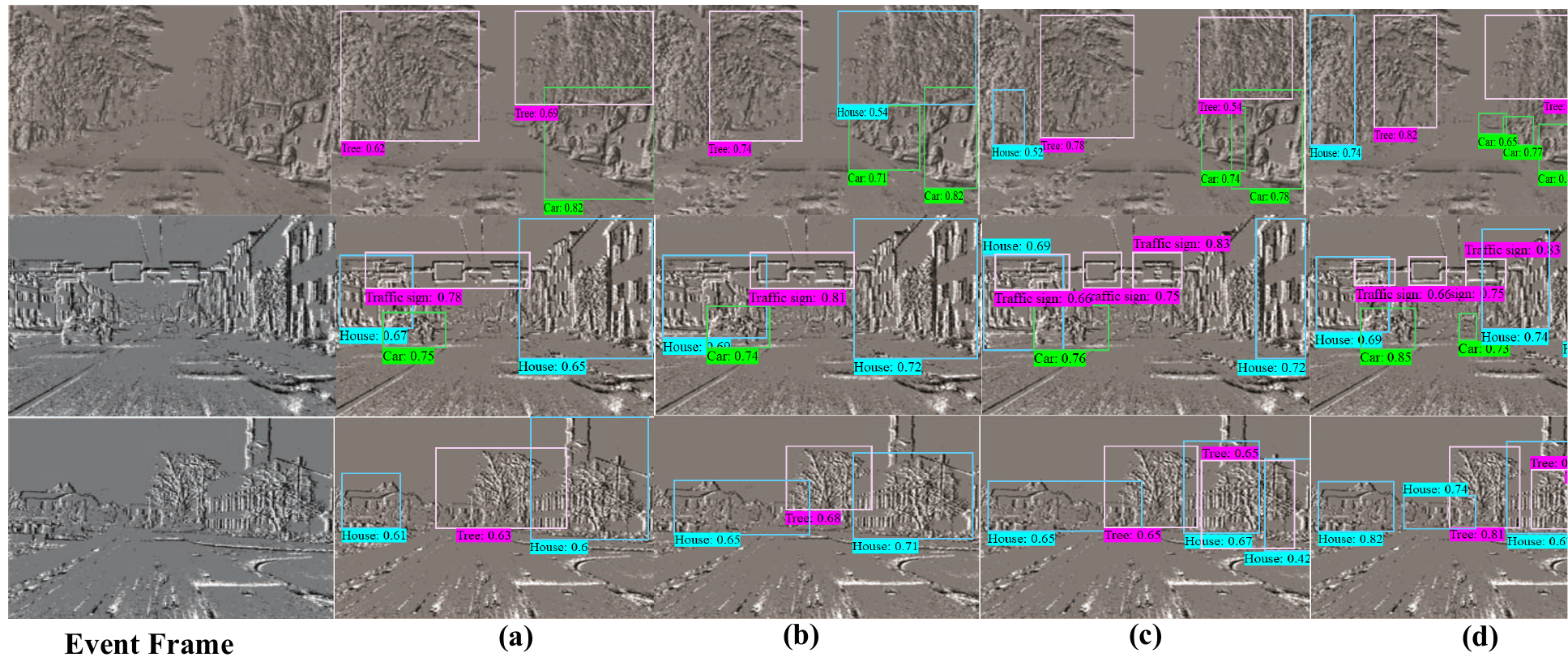}
\end{center}
\vspace{-6mm}
\caption{\textbf{Open Vocabulary Object Detection results on DSEC dataset\cite{gehrig2021dsec}:} From left to right; the models are Event frame; ViLD \cite{gu2021open}; RegionCLIP \cite{zhong2022regionclip}; YOLO-World \cite{cheng2024yolo}; Ours. }
\vspace{-2mm}
\label{openexper}
\end{figure}

\begin{table}[t]
\begin{center}
\resizebox{0.35\textwidth}{!}{
% \begin{tabular}{llllllll}
% \multicolumn{1}{c}{\bf Method}  &\multicolumn{1}{c}{\bf Year} &\multicolumn{1}{c}{\bf AbsRel ↓}   &\multicolumn{1}{c}{\bf RMSE↓} &\multicolumn{1}{c}{\bf $\delta < 1.25$\ ↑}  &\multicolumn{1}{c}{\bf $\delta < 1.25^2$\ ↑}  &\multicolumn{1}{c}{\bf $\delta < 1.25^3$\ ↑} 
\begin{tabular}{cccc}
\hline 
\multicolumn{1}{c}{\bf\ Method}   &\multicolumn{1}{c}{\cellcolor{pink}Architecture}   &\multicolumn{1}{c}{\bf\cellcolor{blue!30} ACC}  
\\ \hline 
HATS \cite{sironi2018hats} &N/A &0.902
\\ \hline 
HybridSNN \cite{kugele2021hybrid} &SNN-CNN&0.906
\\ \hline 
EvS-S \cite{li2021graph} &GNN &0.931
\\ \hline 
HybridSNN \cite{kugele2021hybrid} &SNN &0.770
\\  \hline
Gabor-SNN \cite{sironi2018hats} &SNN  &0.789

\\  \hline
SqueezeNet \cite{cordone2022object} &SNNs &0.846
\\  \hline
MobileNet-64 \cite{cordone2022object} &SNN &0.917
\\  \hline
DenseNet169-16\cite{cordone2022object}  &SNN &0.904
\\  \hline
VGG-11 \cite{cordone2022object} &SNN&0.924 
\\  \hline
SFOD\cite{fan2024sfod} &SNN&0.937
\\  \hline
CREST \cite{mao2024crest} &SNN&0.949
\\  \hline

Ours &SNN+CNN&0.957
\\  \hline
\end{tabular}}
\end{center}
\vspace{-3mm}
\caption{Comparison with existing methods on NCAR dataset \cite{sironi2018hats}. }
\vspace{-9mm}
\label{ncar}
\end{table}

\begin{table}[t]
\begin{center}
\resizebox{0.47\textwidth}{!}{
% \begin{tabular}{llllllll}
% \multicolumn{1}{c}{\bf Method}  &\multicolumn{1}{c}{\bf Year} &\multicolumn{1}{c}{\bf AbsRel ↓}   &\multicolumn{1}{c}{\bf RMSE↓} &\multicolumn{1}{c}{\bf $\delta < 1.25$\ ↑}  &\multicolumn{1}{c}{\bf $\delta < 1.25^2$\ ↑}  &\multicolumn{1}{c}{\bf $\delta < 1.25^3$\ ↑} 
\begin{tabular}{ccccc}
\hline 
\multicolumn{1}{c}{\bf\ Method}  
&\multicolumn{1}{c}{\bf\cellcolor{pink} Car mAP(\%)}  &\multicolumn{1}{c}{\bf\cellcolor{pink}Pedestrian mAP(\%)}  &\multicolumn{1}{c}{\bf\cellcolor{pink}Large vehicle mAP} 
\\  \hline  
 RAMNet \cite{gehrig2021combining}
& 0.244 &0.108 &0.176
\\ \hline 
 SENet \cite{hu2018squeeze}
&0.384 &0.149 &0.260
\\ \hline 
ECANet \cite{wang2020eca} &0.367 &0.128 &0.275
\\  \hline  
CBAM \cite{woo2018cbam}
 &0.377& 0.135& 0.270
\\  \hline  
 SAGate \cite{chen2020bi}
 &0.325 &0.104&0.160
\\  \hline  
 DCF \cite{ji2021calibrated}
 & 0.363 &0.127 &0.280
\\  \hline  
 SPNet \cite{zhou2021specificity}
& 0.392& 0.178 &0.262
\\  \hline  
FPN-Fusion \cite{tomy2022fusing}
 &0.375 &0.109 &0.249
\\  \hline  
DRFuser \cite{munir2023multimodal}
 &0.386 &0.151 &0.306
\\  \hline  
 CMX \cite{zhang2023cmx}
 & 0.416 &0.164 &0.294
\\  \hline  
FAGC \cite{cao2021fusion}
&0.398 &0.144 &0.336
\\  \hline  
RENet \cite{zhou2023rgb}
&0.405 &0.172 &0.306
\\  \hline  
EFNet \cite{sun2022event}
 &0.411 &0.158 &0.326
\\  \hline  
CAFR \cite{cao2024embracing}
 &0.499 &0.258 &0.382
 \\  \hline  
Ours  &0.545 (Basic)&0.312  (Basic) &0.408  (Novel)
 \\  \hline  
\end{tabular}}
\end{center}
\vspace{-3mm}
\caption{ Comparison of the state-of-the-art methods on the DSEC dataset \cite{gehrig2021dsec}. Base Categories: Classes used for training. Novel Category: Unseen class evaluated without training. }
\vspace{-7mm}
\label{DSEC}
\end{table}

\vspace{-3mm}

\subsection{Comparative Study}
\textbf{Novel Category:}
For the DSEC dataset \cite{gehrig2021dsec}, the primary results are shown in Table \ref{DSEC}. All the compared methods are trained using both event streams and image frames. Our model is trained solely on Cars and Pedestrians as base categories, while Large Vehicle is treated as a novel category to evaluate the model's open-vocabulary detection capability. On the base category test set, our model achieves an accuracy of 54.5\% on Car and 31.2\% on Pedestrian. Notably, despite not being trained on Large Vehicle, our model still achieves 42.4\% accuracy, surpassing models that include Large Vehicle in their training data. This result demonstrates the strong generalization ability of our model in open-vocabulary object detection.

% Figure \ref{qualitative} presents a comparison of our model with EFNet \cite{sun2022event}, CAFR \cite{cao2024embracing}, EMS-YOLO \cite{su2023deep}, and event frame-based methods. As shown in the figure, while all models are capable of detecting vehicles, our model achieves the highest accuracy. Furthermore, unlike other approaches, our model can successfully detect novel categories even when trained solely on base categories, whereas the other models fail to achieve this open-vocabulary detection capability. This result further validates the effectiveness of our approach in open-vocabulary object detection for event data, demonstrating superior generalization and adaptability.

\textbf{Zero-shot Object Recognition:}
Our model, trained on the DSEC dataset\cite{gehrig2021dsec}, is directly applied to the NCAR dataset \cite{sironi2018hats} for zero-shot object recognition. Table \ref{ncar} compares our model with other SOTA methods trained on the DSEC dataset. The results demonstrate that our model not only outperforms other SNN-based models in terms of accuracy but also surpasses non-SNN-based models, further validating its effectiveness in zero-shot object recognition tasks.

\textbf{Zero-shot Object Detection:}
We train our model on the DSEC dataset \cite{gehrig2021dsec} and directly apply it to the Gen1 dataset \cite{de2020large} for zero-shot open-vocabulary object detection, with the key results shown in Table \ref{gen1}. Notably, even compared to models trained on Gen1, our model demonstrates superior performance in open-vocabulary detection, highlighting its strong cross-dataset generalization capability.  
Specifically, our model achieves a mean Average Precision (mAP$_{50}$) of 65.7\% at an IoU threshold of 0.5, and mAP$_{50:95}$ of 38.3\% over the IoU range of 0.5 to 0.95. These results validate that our method effectively transfers the knowledge learned from DSEC and enables open-vocabulary detection on unseen event datasets, achieving robust detection performance even for novel categories.

\textbf{Comparison with Open-vocabulary Object Detection: } 
To assess our model’s open-vocabulary performance on event data, gauge its advantage over image-based methods, and evaluate the SNN-driven Adaptive Event Slicing module, we compared it with SOTA open-vocabulary detectors.
We first fed the grayscale event frames provided by the dataset into the image detectors and followed the open-vocabulary protocol: Car and Pedestrian served as base categories for training, while Large Vehicle was kept as a zero-shot novel category. As Table \ref{open} shows, detection accuracy on base classes is markedly higher than on the novel class, confirming that image-trained methods struggle with zero-shot detection on event cameras due to the modality gap between events and images.
Next, we inserted Adaptive Event Slicing as the feature extractor so that features were drawn directly from raw event streams. This raised overall accuracy, demonstrating its effectiveness, but still fell short of our model. Overall, the results indicate that our approach successfully transfers CLIP’s visual knowledge to event data and improves open-vocabulary object detection on event cameras.

Figure \ref{openexper} compares our method with existing open-vocabulary detectors on event data. For a fair evaluation, all methods are tested on the grayscale event frames provided by the dataset. The results indicate that detectors trained on conventional images perform well only when object contours are crisp (e.g., houses, trees) but suffer from mis-detections under occlusion or blur. By contrast, our approach accurately recognises targets even in overlapping or blurred regions, exhibiting greater robustness and generalisation, and confirming its advantage on event-camera inputs.

\subsection {Ablation Study}

\textbf{Adaptive Event Slicing:}
We conducted an ablation study on the two key loss functions in the Adaptive Event Slicing module: Linear Incremental Constraint Loss and Self-Supervised Feedback Loss. As demonstrated in Table \ref{Ablation_result_model}, the adaptive event segmentation strategy significantly improves detection accuracy. By comparing the first and second rows, we observed that Linear Incremental Constraint Loss effectively enhances the model’s object detection accuracy. Furthermore, comparing the second and third rows, we found that Self-Supervised Feedback Loss dynamically adjusts the membrane potential based on object detection results, adaptively optimizing the event segmentation strategy and extracting more discriminative event features, leading to further performance improvement.
\textbf{Knowledge Distillation:}
We investigated the impact of removing the knowledge distillation between image and event data, instead performing object detection using event features extracted by the SNN. As shown in Table \ref{Ablation_result_model} (third and fourth rows), the detection performance drops significantly. This result highlights the critical role of knowledge distillation in feature transfer between event data and images, demonstrating its effectiveness in improving the detector's generalization ability. Building on this, spatial attention was further incorporated. As shown in the fourth and fifth rows, the results indicate that it further improves the effectiveness of knowledge transfer.
\begin{table}[t]
\centering
\resizebox{0.48\textwidth}{!}{
\begin{tabular}{ccccc}
\hline 
\multicolumn{1}{c}{\bf Method} &
\multicolumn{1}{c}{\cellcolor{pink}Architecture} &
\multicolumn{1}{c}{\bf\cellcolor{blue!30}mAP$_{50}$} &
\multicolumn{1}{c}{\bf\cellcolor{blue!30}mAP$_{50:95}$} \\
\hline 
S-Center \cite{bodden2024spiking} & ANN & - & 0.278 \\ \hline
EGO-12 \cite{zubic2023chaos} & ANN & - & 0.504 \\ \hline
Spiking-Yolo \cite{kim2020spiking} & ANN2SNN & - & 0.257 \\ \hline
Spike Calib \cite{li2022spike} & ANN2SNN & 0.454 & - \\ \hline
Spike Transformer v2 \cite{yao2024spike} & ANN2SNN & 0.512 & - \\ \hline
VC-Dense \cite{cordone2022object} & SNN & - & 0.189 \\ \hline
S-Center \cite{bodden2024spiking} & SNN & - & 0.229 \\ \hline
TR-YOLO \cite{yuan2024trainable} & SNN & 0.451 & - \\ \hline
EMS-YOLO \cite{su2023deep} & SNN & 0.501 & 0.301 \\ \hline
SFOD \cite{fan2024sfod} & SNN & 0.5093 & 0.321 \\ \hline
CREST \cite{mao2024crest} & SNN & 0.632 & 0.360 \\ \hline
Ours & SNN + ANN & 0.657 & 0.383 \\ \hline
\end{tabular}}
\vspace{-1mm}
\caption{Comparison of existing methods on the Gen1 dataset \cite{de2020large}.}
\vspace{-8mm}
\label{gen1}
\end{table}
\textbf{Frame Branch:}
We also examined the scenario where the model relies solely on event camera data (without frame camera input). During knowledge distillation, we replaced the frame input of the teacher branch with event reconstruction results \cite{rebecq2019high}. As shown in Table \ref{Ablation_result_model} (fifth and sixth rows), since event-reconstructed images provide only limited visual cues, this substitution generally reduces representation learning quality. Performance degrades compared to knowledge transfer based on real frame data. However, our model remains applicable in cases where no image frames are available.

\begin{table}[t]
\begin{center}
\resizebox{0.47\textwidth}{!}{
% \begin{tabular}{llllllll}
% \multicolumn{1}{c}{\bf Method}  &\multicolumn{1}{c}{\bf Year} &\multicolumn{1}{c}{\bf AbsRel ↓}   &\multicolumn{1}{c}{\bf RMSE↓} &\multicolumn{1}{c}{\bf $\delta < 1.25$\ ↑}  &\multicolumn{1}{c}{\bf $\delta < 1.25^2$\ ↑}  &\multicolumn{1}{c}{\bf $\delta < 1.25^3$\ ↑} 
\begin{tabular}{ccccc}
\hline 
\multicolumn{1}{c}{\bf\ Method}  
&\multicolumn{1}{c}{\bf\cellcolor{pink} Car mAP(\%)}  &\multicolumn{1}{c}{\bf\cellcolor{pink}Pedestrian mAP(\%)}  &\multicolumn{1}{c}{\bf\cellcolor{blue!30}Large vehicle mAP} 

\\ \hline 
&Base  &Base &Novel
\\ \hline 
  ViLD \cite{gu2021open}
&0.343 &0.229 &0.080
\\ \hline 
RegionCLIP \cite{zhong2022regionclip} &0.357 &0.231 &0.085
\\  \hline  
 FVLM \cite{kuo2022f}
 &0.361& 0.235& 0.088
\\  \hline  
  YOLO-World \cite{cheng2024yolo}
 &0.364 &0.237&0.092
\\  \hline  
 Adaptive Event Slicing (SNN) \\ +ViLD \cite{gu2021open}
&0.375 &0.254 &0.092
\\ \hline 
Adaptive Event Slicing (SNN) 
 \\ +RegionCLIP \cite{zhong2022regionclip} &0.381 &0.258 &0.099
\\  \hline  
Adaptive Event Slicing (SNN) \\+ FVLM \cite{kuo2022f}
 &0.392& 0.261& 0.105
\\  \hline  
 Adaptive Event Slicing (SNN) \\+  YOLO-World \cite{cheng2024yolo}
 &0.398 &0.264&0.113
\\  \hline  
Ours  &0.545 &0.312 &0.408  
 \\  \hline  
\end{tabular}}
\end{center}
\vspace{-3mm}
\caption{ Comparison of the state-of-the-art Open-vocabulary Object Detection methods on the DSEC dataset \cite{gehrig2021dsec}. Base Categories: Classes used for training. Novel Category: Unseen class evaluated without training. }
\vspace{-4mm}
\label{open}

\end{table}

\begin{table}[t]
\begin{center}
\resizebox{0.36\textwidth}{!}{
\begin{tabular}{llllll}
\hline 
{\cellcolor{pink}LIC} & 
\multicolumn{1}{c}{\cellcolor{pink}SSF} & 
\multicolumn{1}{c}{\cellcolor{pink}KD} & 
\multicolumn{1}{c}{\cellcolor{pink}Frame} & 
\multicolumn{1}{c}{\cellcolor{blue!30}mAP$_{50}$} & 
\multicolumn{1}{c}{\cellcolor{blue!30}mAP$_{50:95}$} 
\\ \hline 
 &  & &   &0.459 &0.227
\\ \hline 
\checkmark  &   & &  &0.470 &0.232
\\  \hline 
\checkmark    & \checkmark &  &  &0.486  &0.257
\\  \hline 
\checkmark   &\checkmark &  \checkmark  &  &0.621   &0.352  
\\  \hline  
\checkmark   &\checkmark & \checkmark &\checkmark &0.657     & 0.383    
\\  \hline 
% 5   &\checkmark & \checkmark &\checkmark & \checkmark &0.0251    & 0.0437   &  1.565     &0.991
% \\  \hline  
\end{tabular}}
\end{center}
\vspace{-2mm}
\caption{Ablation study of our methods on Gen1:  LIC: Linear Incremental Constraint Loss. SFF: Self-Supervised Feedback Loss. KD: Knowledge Distillation.   }
\vspace{-9mm}
\label{Ablation_result_model}
\end{table}

% \section{ Limitations and Future Work}
% \vspace{-3mm}
% \textbf{Limitations: }
% 1. Low efficiency on GPUs: Running SNNs with dense kernels on GPUs negates the sparsity advantages of event data, leading to redundant MAC operations and high memory consumption.
% 2. Dependence on RGB supervision: The detector currently requires high-quality RGB frames and CLIP-distilled semantics during training; substituting them with event-only reconstructions significantly degrades open-vocabulary detection accuracy.
% \textbf{Future Work:}
% 1. Neuromorphic deployment and energy profiling: Port the network to chips such as Loihi-2 and SpiNNaker to leverage native event-driven sparsity, and systematically quantify the real energy–accuracy gains over GPU inference while identifying hardware bottlenecks for open-vocabulary SNNs.
% 2. Spike-friendly model compression: Combine weight pruning, low-bit quantization, and spike-level sparsification to reduce on-chip storage and inter-layer charge-accumulation overhead.
% 3. self-supervised event training: Develop objectives tailored to asynchronous data—e.g., event–text contrastive alignment or early-stage teacher–student distillation with a frozen RGB detector—and explore lightweight generative modules that recover semantic features directly from events, thereby mitigating the modality gap.

\section{Conclusion} 
\vspace{-2mm}
\label{sec:conclusion}
We present the first open-vocabulary object detection framework for event data, which transfers visual knowledge from CLIP into an event-based detector. To address event stream segmentation, we introduce an SNN-based module that adaptively selects slicing points and extracts discriminative features, forming a collaborative paradigm between SNNs and ANNs.
Experiments demonstrate that our model consistently outperforms dataset-specific baselines. Moreover, even without access to image frames, frame reconstructions from event streams can be used for knowledge distillation, enabling the model to retain both strong learning capacity and robust generalization on event data.

%%%%%%%%%%%%%%%%%%%%%%%%%%%%%%%%%%%%%%%%%%%%%%%%%%%%%%%%%%%%%%%%%%%%%%%%%%%%%%%%

\bibliographystyle{IEEEtran}
\bibliography{egbib}

\end{document}